\documentclass{ieeeaccess}

\usepackage{graphics} 
\usepackage{amsmath} 
\usepackage{booktabs}
\usepackage{cite}
\usepackage{textcomp}
\usepackage{cite}
\usepackage{amsmath,amssymb,amsfonts}
\usepackage{algorithmic}
\usepackage{graphicx}
\usepackage{textcomp}

\usepackage{dsfont}
\usepackage{graphicx}
\usepackage{float}
\usepackage{tabularx}
\usepackage{amsfonts}
\usepackage{multirow}
\usepackage{float}
\usepackage{url}
\usepackage{float} 

\usepackage{color}
\usepackage{soul}

\def\BibTeX{{\rm B\kern-.05em{\sc i\kern-.025em b}\kern-.08em
    T\kern-.1667em\lower.7ex\hbox{E}\kern-.125emX}}

\DeclareMathOperator*{\argmin}{argmin}

\newcommand{\fix}[1]{\textcolor{blue}{#1}}

\begin{document}
\history{Date of publication}
\doi{DOI}
\title{DeepStyle: Multimodal Search Engine\\for Fashion and Interior Design}

\author{\uppercase{Ivona Tautkute}\authorrefmark{1}\authorrefmark{3},
\uppercase{Tomasz Trzcinski}\authorrefmark{2} \authorrefmark{3}\IEEEmembership{Fellow, IEEE},
\uppercase{Aleksander Skorupa}\authorrefmark{3},
\uppercase{Lukasz Brocki}\authorrefmark{1} and
\uppercase{Krzysztof Marasek}\authorrefmark{1}}

\address[1]{Polish-Japanese Academy of Information Technology, Warsaw, Poland (e-mail: s16352 at pjwstk.edu.pl)}
\address[2]{Warsaw University of Technology, Warsaw, Poland (e-mail: t.trzcinski at ii.pw.edu.pl)}
\address[3]{Tooploox, Warsaw, Poland}

\markboth
{Tautkute \headeretal: DeepStyle: Multimodal Search Engine for Fashion and Interior Design}
{Tautkute \headeretal: DeepStyle: Multimodal Search Engine for Fashion and Interior Design}

\corresp{Corresponding author: Ivona Tautkute (e-mail: s16352 at pjwstk.edu.pl).}

\nocite{opencv}

\begin{abstract}
In this paper, we propose a multimodal search engine that combines visual and textual cues to retrieve items from a multimedia database aesthetically similar to the query. The goal of our engine is to enable intuitive retrieval of fashion merchandise such as clothes or furniture. Existing search engines treat textual input only as an additional source of information about the query image and do not correspond to the real-life scenario where the user looks for "the same shirt but of denim".  Our novel method, dubbed DeepStyle, mitigates those shortcomings by using a joint neural network architecture to model contextual dependencies between features of different modalities.
We prove the robustness of this approach on two different challenging datasets of fashion items and furniture where our DeepStyle engine outperforms baseline methods by 18-21\% on the tested datasets.
Our search engine is commercially deployed and available through a ~Web-based~application.
\end{abstract}


\titlepgskip=-15pt

\maketitle

\section{INTRODUCTION}

\IEEEPARstart{M}{ultimodal} search engine allows to retrieve a set of items from a multimedia database according to their similarity to the query in more than one feature spaces, e.g. textual and visual or audiovisual (see Fig.~\ref{semantics}). This problem can be divided into smaller subproblems by using separate solutions for each modality. The advantage of this approach is that both textual and visual search engines have been developed for several decades now and have reached a certain level of maturity. Traditional approaches such as Video Google~\cite{vgoogle} have been improved, adapted and deployed in industry, especially in the ever-growing domain of e-commerce. Major online retailers such as Zalando and ASOS
already offer visual search engine functionalities to help users find products that they want to buy~\cite{asos_zalando}. Furthermore, interactive multimedia search engines are omnipresent in mobile devices and allow for speech, text or visual queries~\cite{interactive-vs,interaction-design, hybrid-vs}.



Nevertheless, using separate search engines per each modality suffers from one significant shortcoming: it prevents the users from specifying a very natural query such as 'I want this type of dress but made of silk'. This is mainly due to the fact that the notion of similarity in separate spaces of different modalities is different than in one multimodal space. Furthermore, modeling this highly dimensional multimodal space requires more complex training strategies and thoroughly annotated datasets. Finally, defining the right balance between the importance of various modalities in the context of a user query is not obvious and hard to estimate a priori. Although several multimodal representations have been proposed in the context of a search for fashion items, they typically focus on using other modalities as an additional source of information, e.g. to increase classification accuracy of compatible and non-compatible outfits~\cite{waseda}.


\Figure[t!](topskip=0pt, botskip=0pt, midskip=0pt)[width=0.4\textwidth]{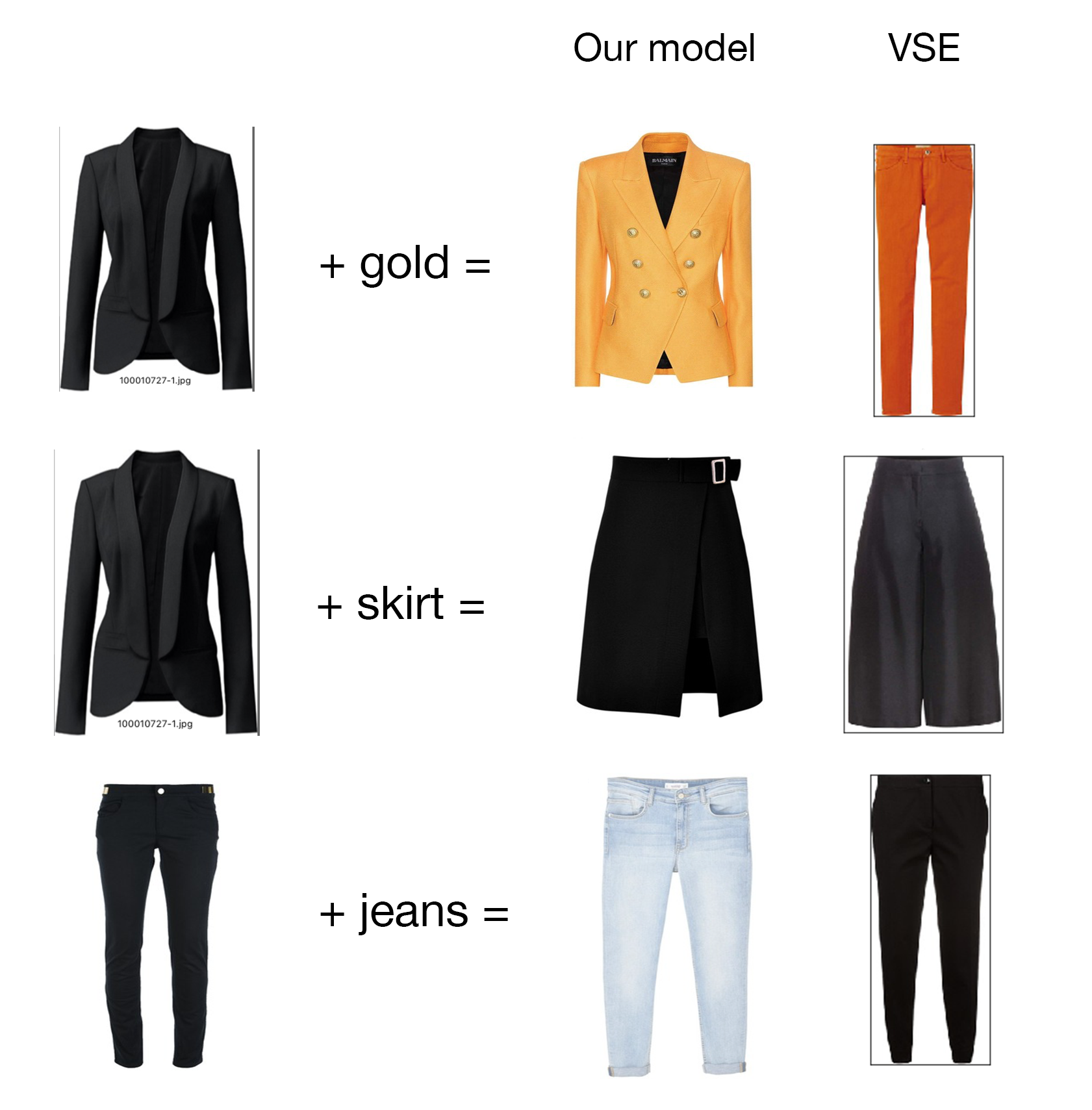} 
{Example of a typical multimodal query sent to a search engine for fashion items. By modeling common multimodal space with a deep neural network, we can provide a more flexible and natural user interface while retrieving results that are semantically correct, as opposed to the results of the search based on the state-of-the-art Visual Search Embedding model \cite{vse}. \label{semantics}}

To address the above-mentioned shortcomings of the currently available search engines, we propose a novel end-to-end method that uses neural network architecture to model the joint multimodal space of database objects. This method is an extension of our previous work \cite{Tautkute17} that blended multimodal results. Although in this paper we focus mostly on the fashion items such as clothes, accessories and furniture, our search engine is in principle agnostic to object types and can be successfully applied in many other multimedia applications. We call our method DeepStyle and show that thanks to its ability to jointly model both visual and textual modalities, it allows for a more intuitive search queries, while providing higher accuracy than the competing approaches. We prove the superiority of our method over single-modality approaches and state-of-the-art multimodal representation using two large-scale datasets of fashion and furniture items. Finally, we deploy our DeepStyle search engine as a web-based application. 

To summarize, the contributions of our paper are threefold:
\begin{itemize}
\item We introduce a novel DeepStyle-Siamese method for retrieval of stylistically similar product items that could be applied to a broad range of domains. To the best of our knowledge, this is the first system for joint learning of stylistic context as well as semantic regularities of both image and text. The proposed method outperforms the baselines on diversified datasets from fashion and interior design domains by 18 and 21\%, respectively.
\item Our system is deployed in production and available through a Web-based application.
\item Last but not least, we introduce a new interior design dataset of furniture items offered by IKEA, an international furniture manufacturer, which contains both visual and textual meta-data of over 2 000 objects from almost 300 rooms. We plan to release the dataset to the public.
\end{itemize} 

The remainder of this work is organized in the following manner. In Sec.~\ref{sec:related} we discuss related work. In Sec.~\ref{sec:style_search_engine} we present a set of methods based on blending single-modality search results that serve as our baseline. Finally, in Sec.~\ref{sec:end-to-end}, we introduce our DeepStyle multimodal approach as well as its extension. In Sec.~\ref{sec:dataset} we present the datasets used for evaluation and in Sec.~\ref{sec:evaluation} we evaluate our method and compare its results against the baseline. Sec.~\ref{sec:conclusions} concludes the paper.

\section{Related Work}
\label{sec:related}
In this section, we first give an overview of the current visual search solutions proposed in the literature. Secondly, we discuss several approaches used in the context of a textual search. We then present works related to defining similarity in the context of aesthetics and style, as it directly pertains to the results obtained using our proposed method. Finally, we present an overview of existing search methods in fashion domain as this topic is gaining popularity. 

\subsection{Visual Search}

Traditionally, image-based search methods drew their inspiration from textual retrieval systems~\cite{nister}. By using $k$-means clustering method in the space of local feature descriptors such as SIFT~\cite{sift}, they are able to mimic textual word entities with the so-called {\it visual words}. Once the mapping from image salient keypoints to visually representative {\it words} was established, typical textual retrieval methods such as Bag-of-Words~\cite{bow} could be used. 
Video Google~\cite{vgoogle} was one of the first visual search engines that relied on this concept. Several extensions of this concept were proposed, {\it e.g.} spatial verification~\cite{philbin} that checks for geometrical correctness of initial query or fine-grained image search \cite{fine-grained} that accounts for semantic attributes of visual words.


Successful applications of deep learning techniques in other computer vision applications have motivated researchers to apply those methods also to visual search. Preliminary results proved that applications of convolutional neural networks~\cite{tolias3} (image-based retrieval), as well as other deep architectures such as Siamese networks~\cite{prod_des} (content-based image retrieval) may be successful, however the concern was raised that they may lack robustness to cropping, scaling and image clutter~\cite{gordo}.

Nevertheless, all of the above-mentioned methods suffer from one important drawback, namely they do not take into account the stylistic similarity of the retrieved objects, which is often a different problem from visual similarity. Items that are similar in style do not necessarily have to be close in visual features space.

\subsection{Textual Search}

First methods that proposed to address textual information retrieval were based on token counts, {\it e.g.} \textit{Bag-of-Words} \cite{bow} or \textit{TF-IDF} \cite{tfidf}. 

Later, a new type of representation called \textit{word2vec} was proposed by Mikolov {\it et. al}~\cite{word2vec}. The proposed models in {\it word2vec} family, namely continuous Bag of Words (CBOW) and Skip-Grams, allow the token representation to be learned based on its local context. To grasp also the global context of the token, GloVe~\cite{glove} has been introduced. GloVe takes advantage of information both from the local context and the global co-occurrence matrix, thus providing a powerful and discriminative representation of textual data. 
Similarly, not all queries can be represented with a text only. There might be a clear textual definition missing for style similarities that are apparent in visual examples. Also, the same concepts might be expressed in synonymical  ways.

\subsection{Stylistic Similarity}
\label{subsec:stylistic_similarity}

Comparing the style similarity of two objects or scenes is one of the challenges that have to be addressed when training a machine learning model for interior design or fashion retrieval application. This problem is far from being solved mainly due to the lack of a clear metric defining how to measure style similarity. Various approaches have been proposed for defining style similarity metric. Some of them focus on evaluating similarity between shapes based on their structures~\cite{kara,kaik} and measuring the differences between scales and orientations of bounding boxes. Other approaches propose the structure-transcending style similarity that accounts for element similarity~\cite{lun}. In this work, we follow \cite{style}, and define style as \textit{a distinctive manner which permits the grouping of works into related categories.} We enforce this definition by including context information that groups different objects together (in terms of clothing items in an outfit or furniture in a room picture in interior design catalog). This allows us to a take data-driven approach that measures style similarity without using hand-crafted features and predefined styles.


\begin{figure*}[t!]
\begin{center}
  \includegraphics[width=0.8\textwidth]{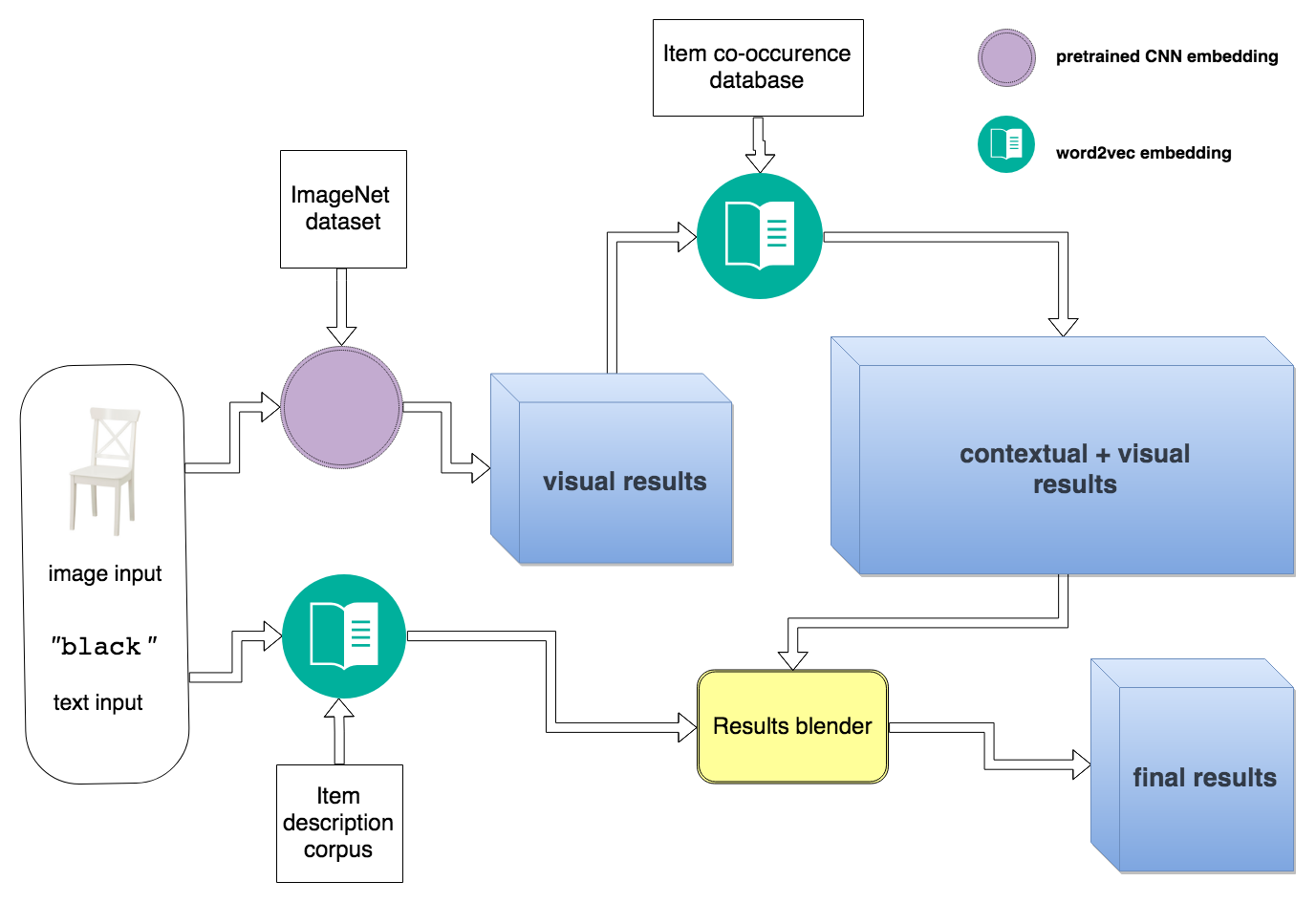} 
    \caption{A high-level overview of our Early-fusion Blending architecture. Visual search finds closest neighbors of the image query in the space of extracted visual features that are the outputs of a pre-trained deep neural network. For each of the retrieved visually similar items, we search for the contextually similar, i.e. items that appeared together in the compatible sets from database, and extend the set of results with those items. The textual block allows to further specify search criteria with text in order to narrow down the set of stylistically and aesthetically similar items, to those that are relevant to the query.}
    \label{pipe}
    \end{center}
\end{figure*}

\subsection{Deep Learning in Fashion}
\label{related_work_fashion}
There has been a significant number of works published in the domain of fashion item retrieval or recommendation due to the potential of their application in highly profitable e-commerce business. Some of them focused on the notion of fashionability, e.g \cite{fashionability} rated a user's photo in terms of how fashionable it is and provided fashion recommendations that would increase overall outfit score. Others focused on fashion items retrieval from online database when presented with user photos taken 'in the wild' usually with phone cameras \cite{street2shop}. Finally, there is ongoing research in terms of clothing cosegmentation \cite{fashion-parsing, clothing-segm} that is an important preprocessing step for better item retrieval results.

Kiros et al. \cite{vse} present an encoder-decoder pipeline that learns a joint Visual-Semantic Embedding (VSE) from images and a text, which is later used to generate text captions for custom images. Their approach is inspired by successes in Neural Machine Translation (NMT) and perceives visual and textual modalities as the same concept described in different languages. The  proposed  architecture  consists  of  LSTM, which is a type of recurrent neural network, for encoding sentences, convolutional neural network (CNN) for encoding images and structure-content neural language  model (SC-NLM) for decoding. The authors show that their learned multimodal embedding space preserves semantic regularities in terms of vector space arithmetic e.g. image of a blue car - "blue" + "red" is near images of red cars. However, results of this task are only available in some example images. We would like to leverage their work and numerically evaluate multimodal query retrieval, specifically in the domain of fashion and interior design.

Ben-younes et al. \cite{mutan} introduced MUTAN, a method for multimodal fusion between visual and textual information using a bilinear framework. It uses a multimodal tensor-based Tucker decomposition in order to efficiently parametrize bilinear interactions between the two representations. Additional low-rank matrix constraint is designed to allow for controlling the full bilinear interaction’s complexity. While in the original paper, authors evaluate architecture primarily on the Visual Question Answering task, we would like to utilize it when learning a joint multimodal representation. In the similar manner, as with the previously mentioned VSE, we evaluate it on multimodal query retrieval in the domain of fashion and interior design.

Xintong Han et al. \cite{polyvore} train bi-LSTM model to predict next item in the outfit generation. Moreover, they learn a joint image-text embedding by regressing image features to their semantic representations aiming to inject attribute and category information as a regularization for training the LSTM. It should be noted, however, that their approach to stylistic compatibility is different from ours in a way that they optimize for generation of a complete outfit (e.g. it should not contain two pairs of shoes) whereas we would like to retrieve items of similar style regardless of the category they belong to. Also, they evaluate compatibility with "fill-in-the-blanks" test that does not incorporate retrieval from the full dataset of items. Only several example results are illustrated and no quantitative evaluation is presented.

Numerous works focus on the task of generating a compatible outfit from available clothing products \cite{waseda,polyvore}. However, none of the related works focus on the notion of multimodality and multimodal fashion retrieval. Text information is only used as an alternative query and not as a complementary information to extend the information about the searched object. Finally, research community has not yet paid much attention to define or evaluate style similarity.

\section{From single to multimodal search}
\label{sec:style_search_engine}

In this section, we present a baseline style search engine model introduced in~\cite{Tautkute17}, which is the basis for our current research. It is built on top of two single-modal modules. More precisely, two searches are run independently for both image and text queries resulting in two initial sets of results. Then, the best matches are selected from initial pool of results according to blending methods - re-ranking  based on visual features similarity to the query image as well as on contextual similarity (items that appear more often together in the same context).

For input, baseline style search engine takes two types of query information: an image containing object(-s), {\it e.g.} a picture of a dining room, and a textual query used to specify search criteria, {\it e.g.} {\it cozy and fluffy}. If needed, an object detection algorithm is run on the uploaded picture to detect objects of classes of interest such as chairs, tables or sofas. Once the objects are detected, their regions of interest are extracted as picture patches and run through visual search method. For queries that already represent a single object, no object detection is required.
Simultaneously, the engine retrieves the results for a textual query. With all visual and textual matches retrieved, our \textit{blending algorithm} ranks them depending on the similarity in the respective feature spaces and returns the resulting list of stylistically and aesthetically similar objects. Below, we describe each part of the engine in more details.

\subsection{Visual Search}

Instead of using an entire image of the interior as a query, our search engine applies an object detection algorithm as a pre-processing step. This way, not only can we retrieve the results with higher precision, as we search only within a limited space of same-class pictures, but we do not need to know the object category beforehand. This is in contrast to other visual search engines proposed in the literature~\cite{prod_des, pinterest}, where the object category is known at test time or inferred from textual tags provided by human labeling. 

\begin{figure}[t!]
\includegraphics[height=140px]{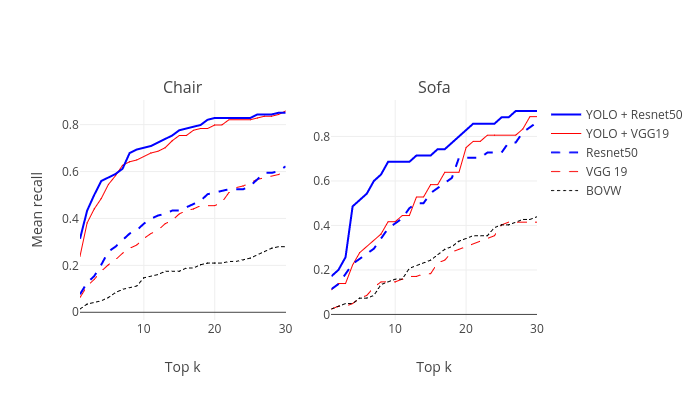} 
\vspace{-0.3cm}
\caption{Architecture comparison for choosing the object detection model in Visual Search. The recall is plotted  as a function of the number of returned items $k$. Best retrieval results are achieved for YOLO object detection and visual features exraction from Resnet-50.}
\label{chair_recall}
\end{figure}

For object detection, we used YOLO 9000~\cite{yolo}, which is based on the DarkNet-19 model~\cite{Darknet, yolo} and is a variety of a neural network. The bounding boxes are then used to generate regions of interest in the pictures and search is performed on the extracted parts of the image.

Once the regions of interest are extracted, we feed them to a pretrained deep neural network to get a vector representation. More precisely, we use the outputs of fully connected layers of neural networks pretrained on ImageNet dataset~\cite{russ}. We then normalize the extracted output vectors, so that their $L_2$ norm is equal to $1$. We search for similar images within the dataset using this representation to retrieve a number of closest vectors (in terms of Euclidean distance). 

To illustrate how the space of extracted visual features preserves the visual similarity of product items, we have visualized the visual features embedding (fig. \ref{w2vec_tsne_embedding_polyvore}) with common dimensionality reduction technique t-SNE \cite{tsne}. It is clearly seen that products that share colour, shape or texture features appear close together.

To determine the pretrained neural network architecture providing the best performance, we conduct several experiments that are illustrated in Fig.~\ref{chair_recall}. As a result, we choose ResNet-50 as our visual feature extraction architecture.

\subsection{Text Query Search}

To extend the functionality of our Style Search Engine, we implement a text query search that allows to further specify the search criteria. This part of our engine is particularly useful when trying to search for product items that represent abstract concepts such as {\it minimalism}, {\it Scandinavian style}, {\it casual} and so on. 

In order to perform such a search, we need to find a mapping from textual information to vector representation of the item, i.e, from the space of textual queries to the space of items in the database. The resulting representation should live in a multidimensional space, where stylistically similar objects reside close to each other.

To obtain the above-defined space embedding, we use a Continuous Bag-of-Words (CBOW) model that belongs to word2vec model family~\cite{word2vec}. In order to train our model, we use the descriptions of items available as a metadata supplied with the catalog images. Such descriptions are available as part of both, the IKEA and the Polyvore datasets, which we describe in details in Sec.~\ref{sec:dataset}. Textual description embedding is calculated as a mean vector of individual words embeddings.

In order to optimize hyper-parameters of CBOW for item embedding, we run a set of initial experiments on the validation dataset and use cluster analysis of the embedding results. We select the parameters that minimize intra-cluster distances at the same maximizing inter-cluster distance. 

Having found such a mapping, we can perform the search by returning $k$-nearest neighbors of the transformed query in the space of product descriptions from the database using cosine similarity as a distance measure.

\subsection{Context Space Search}
\label{context_space}
In order to leverage the information about different item compatibility, which is available as a context data (outfit or room), we train an additional word2vec model (using the CBOW model), where different products are treated as words. Compatible sets of those products appearing in the same context are treated as sentences. 
It is worth noticing that our context embedding is trained without relying on any linguistic knowledge. The only information that the model sees during training is whether given objects appeared in the same set.

\begin{figure}[t!]
\includegraphics[width=0.5\textwidth]{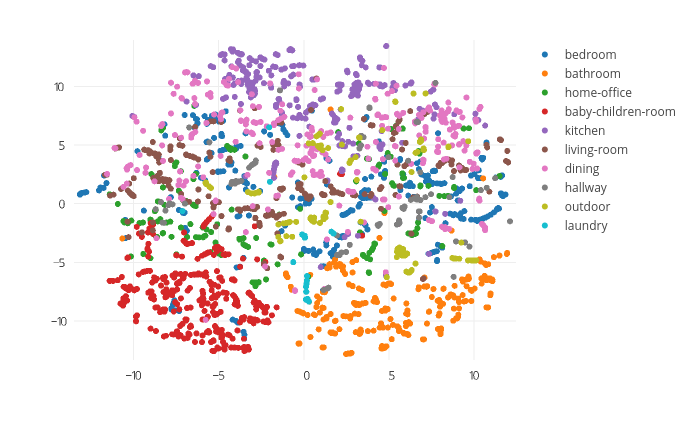} 
\vspace{-0.8cm}
\caption{t-SNE visualization of interior items' embedding using context information only. Distinctive classes of objects, {\it e.g.} those that appear in a bathroom or a baby room, are clustered around the same region of the space. No text descriptions nor information about image room categories was used during training.}
\label{w2vec_tsne_embedding}
\end{figure}

Fig.~\ref{w2vec_tsne_embedding} shows the obtained feature embeddings using t-SNE dimensionality reduction algorithm \cite{tsne} for IKEA dataset. One can see that some classes of objects, {\it e.g.} those that appear in a bathroom or a baby room, are clustered around the same region of the space.


\subsection{Blending Methods}

Let us denote $p= \left( i,t \right)$ to be a representation of a product stored in the database $ \mathbb{P}$. This representation consists of a catalog image $i \in \mathfrak{I}$ and the textual description $ t \in \mathfrak{T}$. The multimodal query provided by the user is given by $Q = (i_q,t_q)$,where $i_q\in \mathfrak{I}$ is the visual query and $ t_q\in \mathfrak{T}$ is the textual query.

We run a series of experiments with blending methods, aiming to combine the retrieval results from various modalities in the most effective way. To that end, we use the following approaches for blending.

\textbf{Late-fusion Blending:}
In the simplest case, we retrieve top $k$ items independently for each modality and take them to as a set of final results. We do not use the contextual information here.

\textbf{Early-fusion Blending:}

In order to use the full potential of our multimodal search engine, we combine the retrieval results of visual, textual as well as contextual search engines in the specific order. We optimize this order to present the most stylistically coherent sets to the user. To that end, we propose \textit{Early-fusion Blending} (see Fig.~\ref{pipe}) approach that uses features extracted from different modalities in a sequential manner.

More precisely, for a multimodal query $(i_q, t_q)$, an initial set of results $R_{vis}$ is returned for visual modality - closest images to $i_q$ in terms of Euclidean distance $d_{vis}$ between their visual representations. Then, we retrieve contextually similar products $R_{cont}$ that are close to $R_{vis}$ results in terms of $d_{cont}$ distance in context embedding space (context space search described in section \ref{context_space}). Finally, $R_{vis}$ and $R_{cont}$ form a list of candidate items $R_{cand}$ from which we select the results  $R$ by extracting the textual features (word2vec vectors) from items descriptors and rank them using distance from the textual query $d_{text}$.

This process can be formulated as: 
\begin{equation}
R_{vis} = \left\lbrace p: \argmin_{p_1,...,p_{n_1} \in \mathbb{P}}{\sum_{j=1}^{n_1} d_{vis}(i_q, i_j)} \right\rbrace 
\Rightarrow
\end{equation}

$$
R_{cont} = \bigcup_{r\in R_{vis}} \left\lbrace p  : \argmin_{p_1,...,p_{n_2} \in \mathbb{P}}{\sum_{j=1}^{n_2} d_{cont}(c_r, c_j)} \right\rbrace \Rightarrow
$$

$$R_{cand} = R_{cont} \cup R_{vis}$$

$$
R = \left\lbrace p: \argmin_{p_1,...,p_{n3} \in R_{cand} }{\sum_{j=1}^{n_3} d_{text}(t_q, t_j )} \right\rbrace 
$$

where $n_1$, $n_2$ and $n_3$ are parameters to be chosen.


\section{DeepStyle: Multimodal Style Search Engine with Deep Learning}
\label{sec:end-to-end}
Inspired by recent advancements in deep learning for computer vision, we experiment with end-to-end approaches that learn the embedding space jointly.
In this section, we describe experiments with artificial neural networks that we did to create a joint image-text model. Our goal is to have one model that takes image and text and returns product items satisfying both modalities. First, we start with a simple approach and experiment with a single neural network that is fed with multiple inputs and learns a multimodal embedding space. Such embedding can later be used to retrieve results using a multimodal query. The first proposed architecture is a multimodal DeepStyle network that learns common image-text embedding through classification task. Then, we go further and improve over the first network with the information we have about products' context (outfit). The most straightforward way to make neural network learn the distances between similar and non-similar items is by introducing a Siamese architecture with shared weights and contrastive loss. The resulting architecture that learns to map pairs from the same outfit close in the multi-modal embedding space is called DeepStyle-Siamese network.

\textbf{DeepStyle: }Our proposed neural network learns common embedding through classification task. Our architecture, dubbed \textit{DeepStyle}, is inspired by~\cite{waseda}, where they use a multimodal joint embedding for fashion product retrieval. In contrast to their work, our goal is not to retrieve images with text query (or vice versa) but to retrieve items where a text query compliments the image and provides additional query requirements.

\begin{figure*}[t!]
\centering
\includegraphics[width=0.9\textwidth]{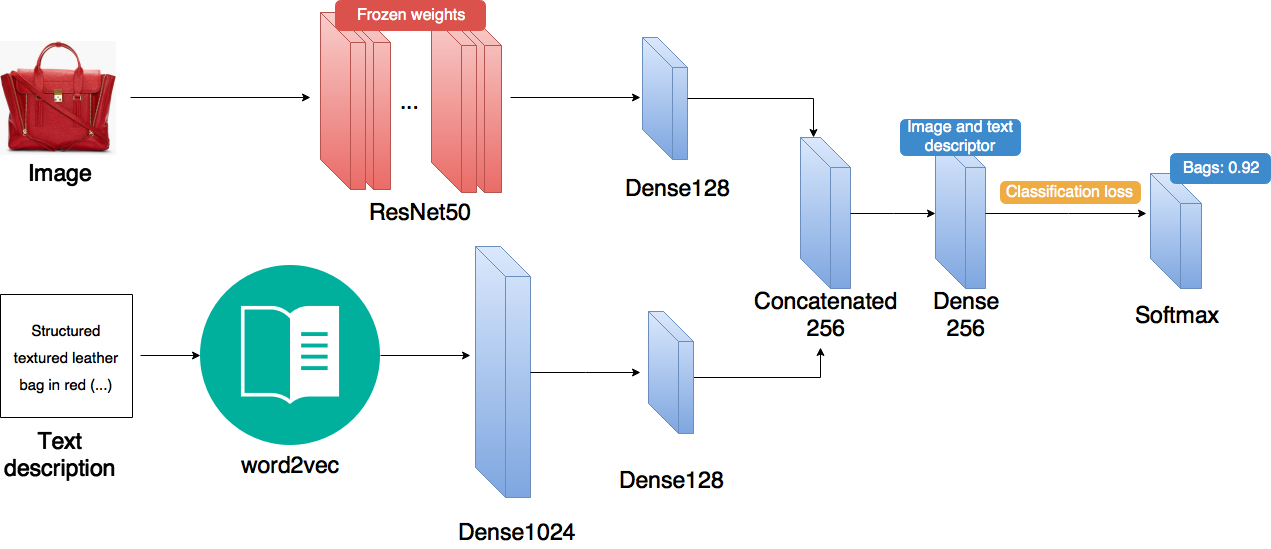} 
\caption{The proposed architecture of DeepStyle network. An image is first fed through the ResNet-50 network pretrained on ImageNet, while the corresponding text description is transformed with word2vec. Both branches are compressed to 128 dimensions, then concatenated to common vector. Final layer predicts a clothing item category. Penultimate layer serves as a multimodal image and text representation of a product item.}
\label{architecture_m}
\end{figure*}

Similarly to \cite{waseda}, our network has two inputs - image features (output of penultimate layer of pretrained CNN) and text features (processed with the same \textit{word2vec} model trained on descriptions). We then optimize for classification loss to enforce the concept of semantic regularities. For this purpose, product category labels (with arbitrary number of classes) should be present in the dataset. Unlike \cite{waseda}, we do not consider the image and the text branches separately for predictions but add a fully connected layer on top of the concatenated image and text embeddings that is used to predict a single class. Illustration of network architecture is presented in fig.~\ref{architecture_m}.

\begin{figure}[t!]
\includegraphics[width=0.5\textwidth]{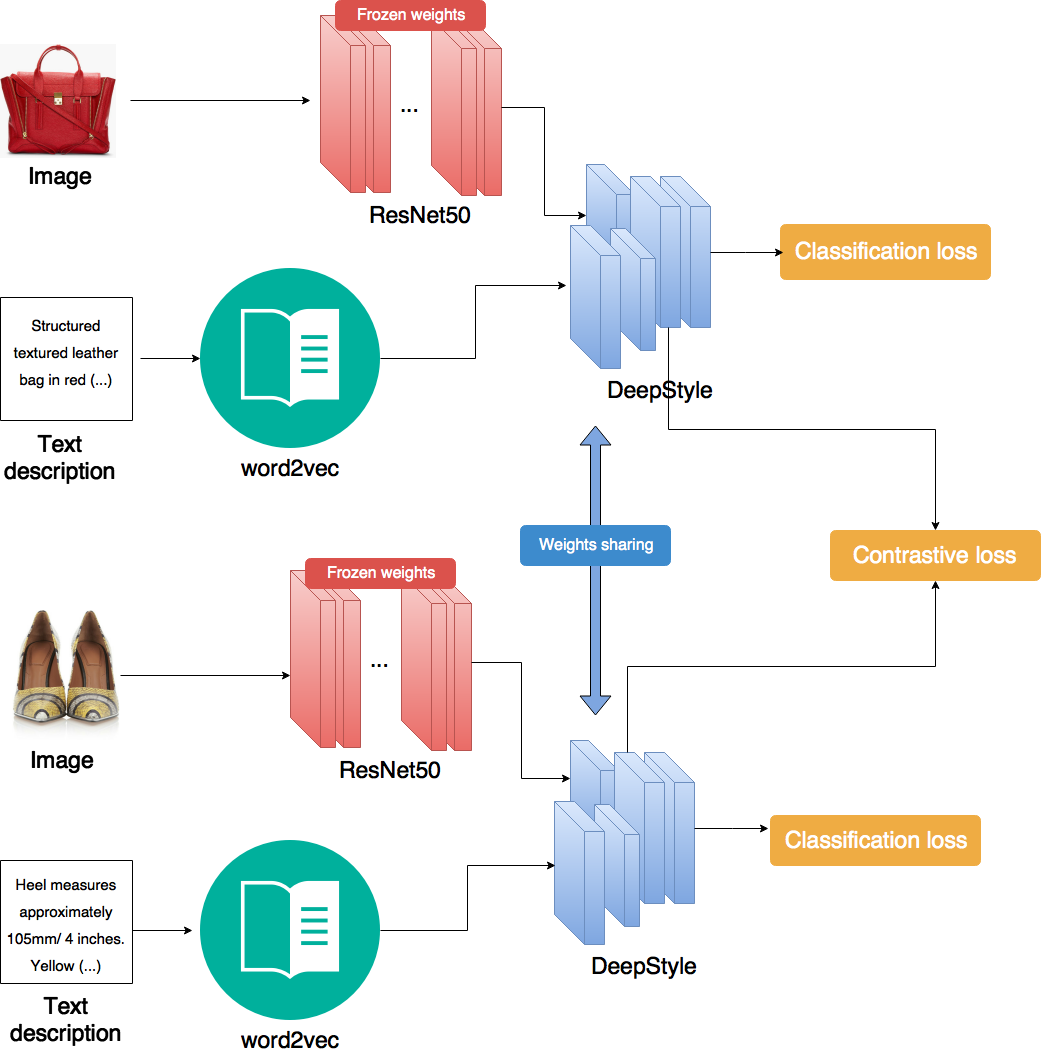} 
\caption{The architecture of DeepStyle-Siamese network. DeepStyle block is the block of dense and concatenation layers from Fig. \ref{architecture_m} that has shared weights between the image-text pairs. Three kinds of losses are optimised - the classification loss for each image-text branch and the contrastive loss for image-text pairs. Contrastive loss is computed on joint image and text descriptors.}
\label{architecture_s}
\end{figure}

\textbf{DeepStyle-Siamese:}
We want to also include context information (whether or not two items appeared in the same context) to our network. For this purpose, we design a Siamese network \cite{siamese} where each branch has a dual input consisting of image and text features. Positive pairs are generated as image-text pairs from the same outfit while unrelated pairs are obtained by randomly sampling an item (image and description) from a different outfit.

As seen in fig.~\ref{architecture_s}, two types of losses are optimized. Classification loss is used as before to help network learn semantic regularities. Also, minimizing contrastive loss encourages image-text pairs from the same outfit to have a small distance between embedding vectors while different outfit items to have distance larger than a predefined margin.

Formally, contrastive loss is defined in the following manner \cite{siamese}:
\begin{equation}
L_{C}(d, y) = (1-y)\frac{1}{2}d^2+ y \frac{1}{2}\{max(0, m-d)\}^2,
\end{equation}
where $d$ is the Euclidean distance between two different embedded image-text vectors $(i,t)$ and $(i',t')$, $y$ is a binary label indicating whether two vectors are from the same outfit ($y=0$) or from different outfits ($y=1$) and $m$ is a predefined margin for the minimal distance between items from different outfits.

Full training loss $L$ consists of weighted sum of contrastive loss and cross entropy classification losses:

\begin{equation}
L = \alpha L_{C}(d,y) + \beta L_{X}(Cl_1(i, t), \tilde{y}(i, t)) + 
\end{equation}
$$+ \gamma L_{X}(Cl_2(i',t') , \tilde{y}(i', t')), $$

where $L_X$ is the cross entropy loss, $Cl_1(i,t)$ and $Cl_2(i,t)$ are outputs of the first and second classification branches respectively and $\tilde{y}(i,t)$ is the category label for product with image $i$ and text description $t$. Parameters $\alpha, \beta, \gamma$ are treated as hyperparameters for tuning.

\begin{figure}[t]
  \begin{center}
    \includegraphics[width=0.5\textwidth]{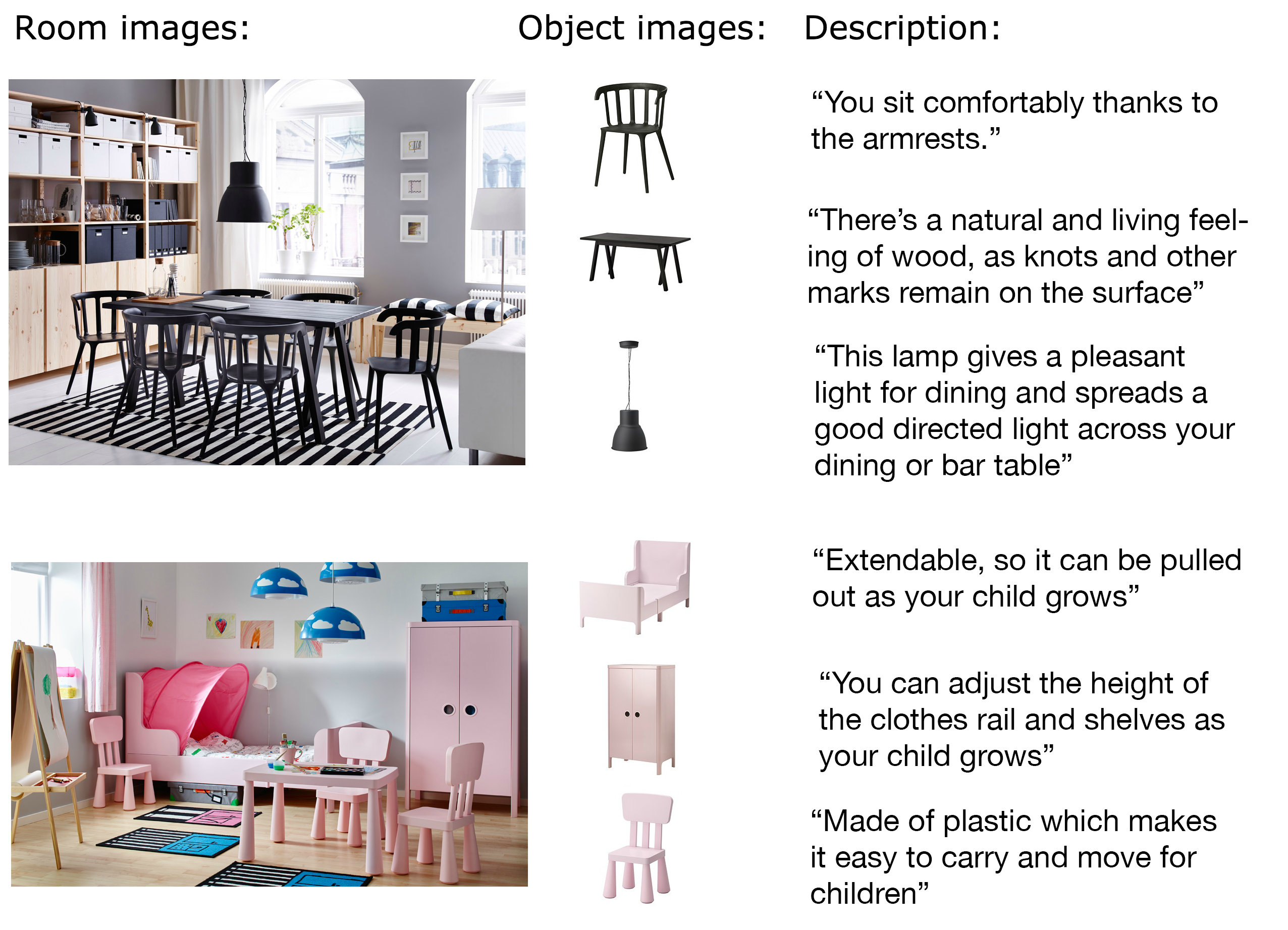} 
    \caption{Example entries from IKEA dataset. It contains room images, object images and their respective text descriptions.}
    \label{ikea_dataset_description}
  \end{center}
\end{figure}

\section{Datasets}
\label{sec:dataset}

Although several datasets for standard visual search methods exist, {\it e.g.} Oxford 5K \cite{philbin} or Paris 6K \cite{philbin2}, they are not suitable for our experiments, as our multimodal approach requires an additional type of information to be evaluated. More precisely, dataset that can be used with a multimodal search engine should fulfill the following conditions:
 
\begin{itemize}
\item It should contain both images of individual objects as well as scene images (room/outfit image) with those objects present.
\item It should have a ground truth defining which objects are present in scene photo.
\item It should also have textual descriptions.
\end{itemize}

We specifically focus on datasets containing pictures of interior design and fashion as both domains are highly dependant on style and would benefit from style search engine applications. In addition, we analyze datasets with varying degrees of context information, as in real life applications it might differ from dataset to dataset. For example, in some cases (specifically when the database is not very extensive), items can co-occur very often together (in context of the same design, look or outfit). Whereas in other cases, when database of available items is much bigger, the majority of items will not have many co-ocurrences with other items. We apply our Multimodal Search Engine for both types of datasets and perform quantitative evaluation to find the best model.


\subsection{Interior Design}

To our knowledge, there is no publicly available dataset that contains the interior design items and fulfill previously mentioned criteria. Hence, we collect our own dataset by scraping the website of one of the most popular interior design distributors - IKEA\footnote{\textit{https://ikea.com/}}. We collect 298 room photos with their description and 2193 individual product photos with their textual descriptions. A sample image of the room scene and interior item along with their description can be seen in Fig.~\ref{ikea_dataset_description}. We also group together products from some of the most frequent object classes ({\it e.g.} chair, table, sofa) for more detailed analysis. In addition, we divide room scene photos into 10 categories based on the room class (kitchen, living room, bedroom, children room, office). 
The vast majority of furniture items in the dataset (especially from the frequent classes above) have rich context as they appear in more than one room.


\subsection{Fashion}

\begin{figure}[t]
  \begin{center}
    \includegraphics[width=0.5\textwidth]{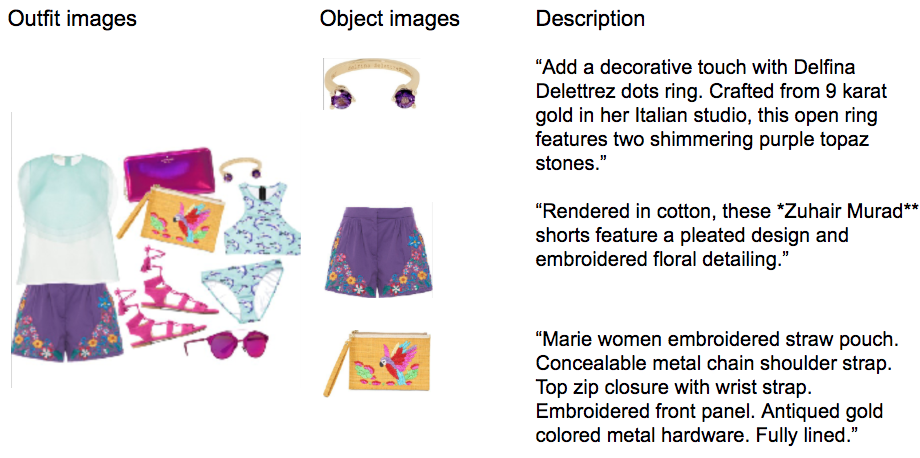} 
    \caption{Example entries from Polyvore dataset. It contains outfit images, item images and their respective text descriptions.}
    \label{examplify_polyvore}
  \end{center}
\end{figure}

\begin{figure}[b]
  \begin{center}
    \includegraphics[width=0.5\textwidth]{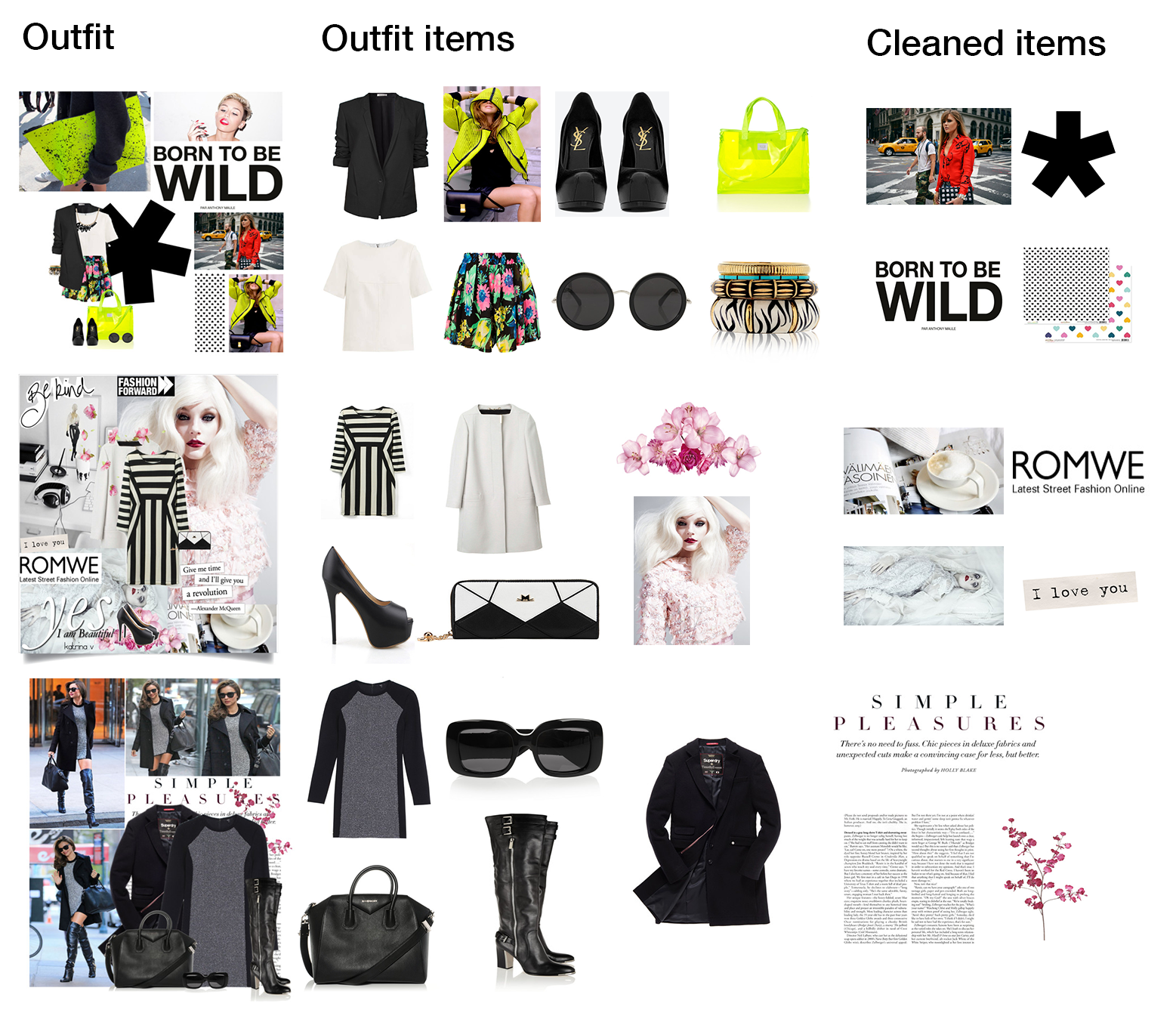} 
    \caption{Illustration of Polyvore dataset. Cleaned items column shows items removed from the original dataset \cite{polyvore} after cleaning procedures.}
    \label{polyvore_dataset}
  \end{center}
\end{figure}

Several datasets for fashion related tasks are already publicly available. \textit{DeepFashion} \cite{deepfashion} contains ~800~000 images divided into several subsets for different computer vision tasks. However, it lacks the context (outfit) information as well as the detailed text description. \textit{Fashion Icon} \cite{fashion-parsing} dataset contains video frames for human parsing but no individual product images. In contrast, \textit{Polyvore} \cite{polyvore} dataset has satisfied our dataset conditions mentioned before (see Fig.~\ref{examplify_polyvore}).

Polyvore dataset contains 111~589 clothing items that are grouped into compatible outfits (of 5-10 items per outfit). We perform additional dataset cleaning - remove non-clothing items such as electronic gadgets, furniture, cosmetics, designer logos, plants, furniture. In addition, we perform additional scraping of Polyvore\footnote{\textit{http://polyvore.com}} website for product items in the cleaned dataset to obtain longer product descriptions and add the descriptions where they are missing.
As a result, we have 82~229 items from 85 categories with text descriptions and context information. Context information is much weaker when compared to IKEA dataset. Only $30\%$ of clothing items appear in more than one outfit. 
Sample images from Polyvore dataset together with illustration of cleaning procedure are shown in fig. \ref{polyvore_dataset}.

Item (query) images are already object photos. Therefore, for fashion dataset object detection step from style search engine is omitted for evaluation. 


\begin{figure}[t!]
\includegraphics[width=0.5\textwidth]{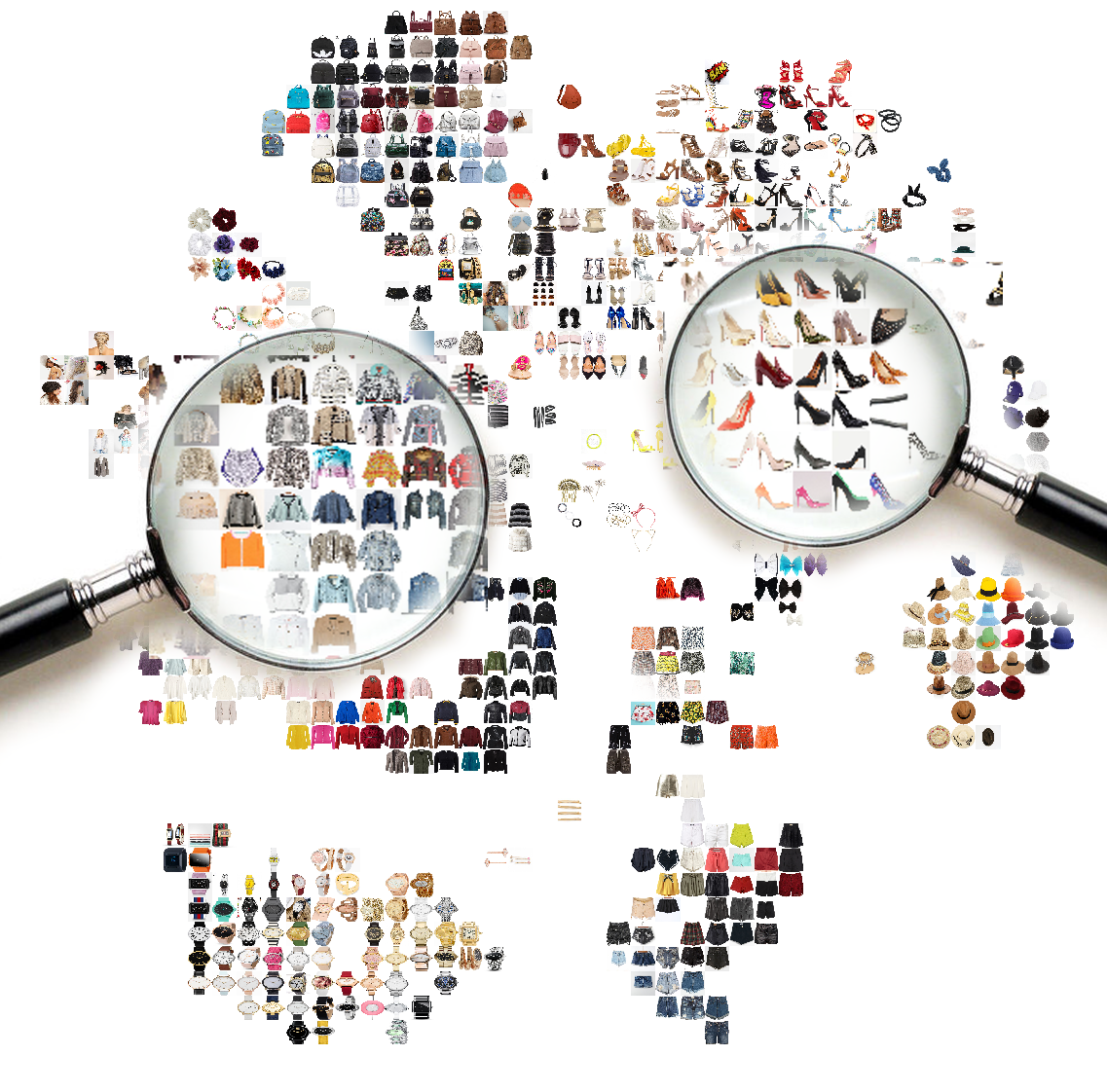} 
\vspace{-0.8cm}
\caption{t-SNE visualization of clothing items' visual features embedding. Distinctive classes of objects, {\it e.g.} those that share visual similarities are clustered around the same region of the space.}
\label{w2vec_tsne_embedding_polyvore}
\end{figure}


\section{EVALUATION}
\label{sec:evaluation}
In this section we want to evaluate our method and check how well it performs in the task of finding similar items when compared to baselines by querying on a subset of images and a set of popular text queries.
\subsection{Evaluation Metrics}
{\bf Similarity score:} 
As mentioned in Sec.~\ref{subsec:stylistic_similarity}, defining a similarity metric that allows quantifying the stylistic similarity between products is a challenging task and an active area of research. In this work, we propose the following similarity measure that is inspired by~\cite{style} and based on the probabilistic data-driven approach.

Let us remind that $\mathcal{P}$ is a set of all possible product items available in the catalog.
Let us then denote $\mathcal{C}$ to be a set of all sets that contain stylistically compatible items (such as outfits or interior design rooms).
Then we search for a similarity function between two items $p_1$, $p_2 \in \mathcal{P}$ which determines if they fit well together. We propose the empirical similarity function $s_c:  \mathcal{P} \times  \mathcal{P} \rightarrow \left[ 0,1 \right]$ which is computed in the following way:
\begin{equation}
	s_{c}(p_1, p_2) = \frac{| \{ C_{i} \in \mathcal{C}: p_1 \in C_{i} \wedge  p_2 \in C_{i} \}|}{ \max_{ p \in \{p_1, p_2\} } \; \mid \{ C_{j} \in \mathcal{C} : p \in  C_{j} \} \mid}.
\end{equation}
In fact, it is the number of compatible sets $\mathcal{C}_{i}$ that are empirically found from  $\mathcal{C}$, in which both  $p_1$ and $p_2$ appear, normalized by the maximum number of compatible sets in which any of those items occur. This metric can be interpreted as an empirical probability for the two objects $p_1$ and $p_2$ to appear in the same compatible set and it is expressed by the similarity score lying in the interval $\left[ 0,1 \right]$

In order to account for datasets that have weak context information (where two items rarely co-occur in the same compatible set), we add an additional similarity measure $s_n$ that is directly derived from their name overlap. It counts for overlap of some of the most frequent descriptive words such as \textit{elegant, denim, casual, etc}. It should be mentioned, however, that product name information should be independent from the text description (that is used during training). As a result, name-derived similarity is non-zero only on datasets that have this kind of additional name information.

\begin{equation}
s_n(p_1, p_2) = \mathds{1}\{W_{p_1} \cap W_{p_2} \neq \emptyset\},
\end{equation}
where $W_f$ is a set of frequent descriptive words appearing in the name of item $f$.


To summarize, an evaluated pair is considered to be similar if either of the two conditions is satisfied:
\begin{itemize}
    \item items co-occurred in the same outfit before
    \item names of the two items are overlapping
\end{itemize}

Formally,
\begin{equation}
s(p_1, p_2) = \max \left( s_c(p_1, p_2), s_n(p_1, p_2) \right) .
\end{equation}


{\bf Intra-List similarity:}
Given that our multimodal query search engine provides a non-ranked list of stylistically similar items, the definition of the evaluation problem differs significantly from other information retrieval domains. For this reason, instead of using some of the usual metrics for performance evaluation like mAP~\cite{MAP} or nDCG~\cite{cumulatedgain}, which use a ranked list of items as an input, we apply a modified version of the established metric for non-ranked list retrieval. Inspired by the~\cite{intra_list_similarity}, we define the average intra-list
similarity for a generated results list  $R$ of length $k$ to be:
\begin{equation}
	AILS(R) =  {{k}\choose{2}}^{-1} \sum_{p_i \in R} \sum_{p_j \in R , p_i \neq p_j}	s(p_1, p_2),
\end{equation}
that is an average similarity score computed across all possible pairs in the list of generated items. By doing so, we are aiming to assess the overall compatibility of the generated set. As mentioned in~\cite{intra_list_similarity}, this metric is also permutation-insensitive, hence the order of retrieved results does not matter, making it suitable for not ranked results.    

\subsection{Baseline}

In experiments, we compare our approach with several baselines.

Recent multimodal approach that could be used for item retrieval is Visual Search Embedding (VSE) \cite{vse}. For evaluation, we fine-tune the weights of a pretrained model made publicly available by authors on our datasets. The model was pretrained on MS COCO dataset that has 80 categories with broad semantic context, hence it's  applicable to our datasets. Original VSE implementation uses VGG~19 \cite{vgg} architecture for feature extraction. Our model extracts deep features with Resnet~50, architecture that has more layers and parameters. Hence, to allow a fair comparison, we train an additional baseline model with VSE that also uses Resnet-50 as a feature extractor. For coherence, we also include a version of our model that uses VGG-19 as feature extractor.

Another approach for multimodal representation learning from text and image is MUTAN \cite{mutan}. It is a multimodal tensor-based Tucker decomposition that efficiently parametrizes bilinear interactions between visual and textual representations. We take an intermediate representation of MUTAN model that is a fusion of image and text and use it as feature extractor for products database.

We also compare our method with Late and Early-fusion Blending strategies.

\subsection{Results}

{\textbf{Evaluation protocol:}}
In order to test the ability of our method to generalize, we evaluate it using a dataset different from the training dataset. For both datasets, we set aside 10\% of the initial number of items for that purpose. All results shown in this section come from the following evaluation procedure:
\begin{enumerate}
\item For each item/text query from the test set we extract visual and textual features.
\item We run engine and retrieve a set of $k$ most compatible items from the trained embedding space.
\item We evaluate the query results by computing an Average Intra-List Similarity metric for all possible pairs between the retrieved items and the query, which gives ${{k}\choose{2}}$ pairs for $k$ retrieved items.
\item  The final results are computed as the mean of AILS scores for all of the tested queries.
\end{enumerate}
It should be noted that for the IKEA dataset, object detection is performed on room images and similar items are returned for the most confident item in the picture. On the other hand, for Polyvore dataset, the test set images are already catalog items of clothes on white background, hence the object detection is not necessary and this step is omitted.

\begin{table*}[t!]
\setlength{\tabcolsep}{10pt}
\def\arraystretch{1.1}
\centering
\caption{Mean AILS results averaged for IKEA dataset and sample text queries from the set of most frequent words in text descriptions. }
\label{results_similarity_ikea}

\begin{tabular}{@{}cccc|cc|ccc@{}}
\toprule
\multirow{2}{*}{Text query} &  \multirow{2}{*}{MUTAN~\cite{mutan}} &  \multirow{2}{*}{{VSE-Resnet~\cite{vse}}} & \multirow{2}{*}{VSE-VGG19~\cite{vse}}  & \multicolumn{2}{c|}{Blending \cite{Tautkute17}}   & \multirow{2}{*}{DeepStyle} & \multicolumn{2}{c}{DeepStyle-Siamese}    \\
& & & & Late-fusion & Early-fusion &  & VGG19 & Resnet\\
\midrule

\textit{decorative}  & 0.1589 & 0.1526 & 0.1475  & 0.2742 &0.2332 &0.2453 & 0.2526 & \textbf{0.2840}  \\ 
 \textit{black}	& \textbf{0.3271} & 0.1928 & 0.3217& 0.2361 & 0.2354 & 0.1967  & 0.2200 & 0.2237\\ 
\textit{white}	 &  0.1588 & 0.1693	& 0.1476 & 0.2534 &0.2048 & 0.1730 & 0.2534 & \textbf{0.2742} \\ 
 \textit{smooth}  & 0.0011 & 0.1158 & 0.1648 & 0.2667 &0.2472& \textbf{0.3022} & 0.2667 & 0.2642\\ 
\textit{cosy} & 0.1121 & 0.2116 & 0.2918  & 0.1073  & 0.2283& \textbf{0.3591} & 0.2730& 0.2730\\ 
\textit{fabric} & 0.0280 & 0.2437 & 0.1038	& 0.1352  & 0.2225	& 0.0817 & 0.2225 & \textbf{0.2487}\\ 
\textit{colourful}	 & 0.1121 & 0.1839	& 0.3163  & 0.2698 & 0.2327& \textbf{0.3568}&  0.2623 & 0.2623\\ \midrule
Average  &	 0.1295 & 0.1814 & 0.2134  & 0.2164&  0.2287& 0.2449 & 0.2501& \textbf{0.2589}\\
\bottomrule
\end{tabular}
\end{table*}



\begin{table*}[t!]
\setlength{\tabcolsep}{10pt}
\def\arraystretch{1.1}
\centering
\caption{Mean AILS results for Fashion Search on Polyvore dataset. Sample text queries are selected from the set of most frequent words in text descriptions}
\label{tab:results_similarity_poly}
\begin{tabular}{@{}cccc|cc|ccc@{}}
\toprule
\multirow{2}{*}{Text query} & \multirow{2}{*}{MUTAN~\cite{mutan}} &  \multirow{2}{*}{VSE-Resnet~\cite{vse}} & \multirow{2}{*}{VSE-VGG19~\cite{vse}} & \multicolumn{2}{c|}{Blending \cite{Tautkute17}}   & \multirow{2}{*}{DeepStyle} & \multicolumn{2}{c}{DeepStyle-Siamese} \\
& & & & Late-fusion & Early-fusion &  & VGG19 & Resnet \\
\midrule
 \textit{black} & 0.1520 & 0.2580 & 0.2932 & 0.2038 & 0.3038 & 0.2835 & \textbf{0.3532} & 0.2719 \\
 \textit{white} & 0.1682  & 0.2610 & 0.2524 & 0.2047 & 0.2898 & 0.2012 & \textbf{0.3499} &  0.2179 \\
 \textit{leather} & 0.1510 & 0.2607 & 0.2885  & 0.2355 & 0.2946 & 0.2510 &0.3079 & \textbf{0.3155} \\
 \textit{jeans} & 0.1668 & 0.2565 &  0.2381 & 0.1925 & 0.2843 & \textbf{0.4341} & 0.3804 & 0.4066 \\
 \textit{wool} & 0.1603 & 0.2578 & 0.3025  & 0.1836 & 0.2657 & \textbf{0.5457} & 0.3458 & 0.4337 \\
 \textit{women} & 0.1491 & 0.2566 & 0.2488  & 0.1931 & 0.3088 & \textbf{0.3808} & 0.2249 & 0.3460 \\
 \textit{men} &  0.1910 & 0.2662 & 0.2836  & 0.1944 & \textbf{0.2900} & 0.1961 & 0.2200 & 0.2549 \\
 \textit{floral} & 0.1642 & 0.2635 & 0.2729 & 0.3212 & 0.2954 & \textbf{0.3384} & 0.2521 & 0.2858 \\
 \textit{vintage} & 0.1782 & 0.2567  & 0.2986  & 0.3104 & 0.3035 & 0.3317 & 0.3286 & \textbf{0.3935} \\
 \textit{boho} &  0.1577 & 0.2597 & 0.2543 & 0.3074 & 0.2893 & 0.2750 & 0.3162 & \textbf{0.3641} \\
 \textit{casual} & 0.1663 & 0.2626 & 0.2808  & 0.3361 & 0.3030 & 0.2071 &\textbf{0.4306} & 0.2693 \\ 

\midrule
\textbf{Average} & 0.1504 & 0.2383 & 0.2740 & 0.2439 & 0.2935 & 0.3131 & 0.3191 & \textbf{0.3236}\\
\bottomrule
\end{tabular}
\end{table*} 

{\textbf{Quantitative results:}}
Tab.~\ref{results_similarity_ikea} shows the results of the blending methods for the IKEA dataset in terms of the mean value of our similarity metric. 



When analyzing the results of blending approaches, we experiment with several textual queries in order to evaluate system robustness towards changes in the text search.
We observe that DeepStyle approach outperforms all baselines for almost all text queries achieving the highest average similarity score. DeepStyle-Siamese approach gives the best results, outperforming the strongest baseline (VSE-VGG19) by $21\%$ for IKEA dataset. It should also be noted that network complexity is not directly correlated with its ability to learn style similarity that is illustrated by worse similarity results on VSE baseline that extracts Resnet-50 features instead of VGG-19. For coherence, we include an additional experiment of training DeepStyle-Siamese network with VGG-19 feature extraction as input. Similarity values on test set for this DeepStyle version are slightly worse than trained with Resnet features, however the difference is not significant.

\begin{figure}[t!]
\label{bar_chart}
\includegraphics[width=0.5\textwidth]{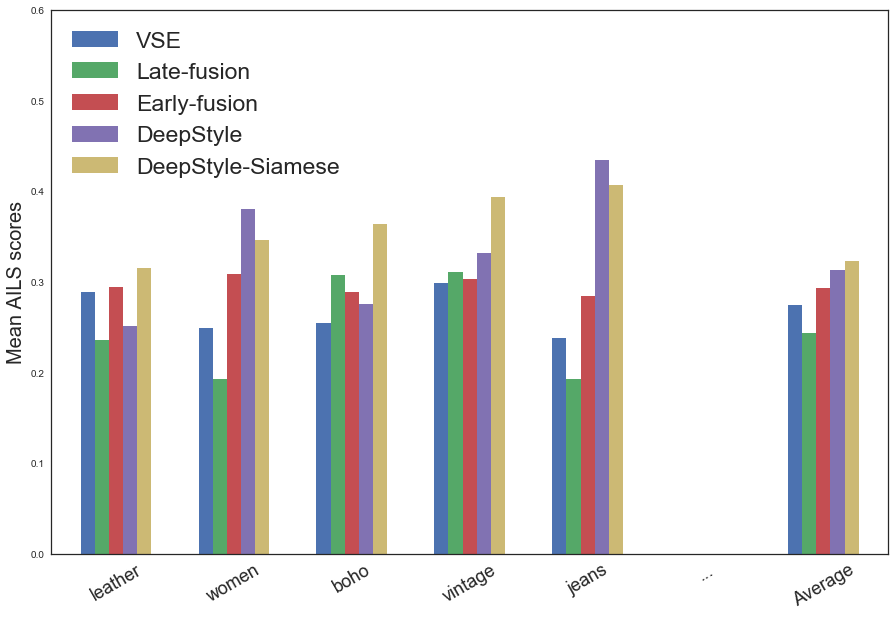} 
\begin{center}
\centering
\caption{Mean AILS metric scores for selected textual queries and the average of the mean scores for other methods and strongest baseline (VSE-VGG). We can see that our DeepStyle-Siamese architecture significantly outperforms other architectures on multiple text queries.}
\end{center}
\vspace{0.6cm}
\end{figure}

Tab.~\ref{tab:results_similarity_poly} shows the results of all of the tested methods for the Polyvore dataset in terms of the mean value of our similarity metric. 
Here, we also evaluate two joint architectures, namely DeepStyle and DeepStyle-Siamese. Fig.\ref{bar_chart} shows that DeepStyle architecture yields better results in terms of an average performance over different textual queries, when compared to our previous blending approaches, as well as other baselines. In this case, DeepStyle-Siamese also yields the best average similarity results. In terms of an average performance, it scores by 32\% higher, when compared to the strongest baseline model,  and more than 4\% higher, when compared to DeepStyle.

It can be observed by the reader, that adding contextual information helps both systems to achieve better results. For blending approaches, Early-fusion, where the contextual embedding was used as a part of the retrieval process, outperforms Late-fusion, where this embedding was not used. Similarly, DeepStyle-Siamese architecture which was learned using matching pairs of furniture, hence implicitly using contextual information, outperforms plain DeepStyle architecture which was not using it.

\textbf{Text query analysis:} The choice of text queries for input is completely arbitrary as they provide additional description that does not have to be related to image content. Hence we analyze if any types of text queries work better with our model. We group the set of most common descriptive words in Polyvore descriptions by separate categories, such as fabrics (leather, suede, denim), color, style (floral, vintage, classic) and human body (ankle, skinny, average). The comprehensive analysis is presented in table \ref{similarity_per_category}. One may observe that 
text queries related to color give slightly better similarity results. This might seem intuitive as the concept of color seems easy to define and learn. On the other hand, the average similarity difference is not significant between various text groups, implying that all types of queries can be used with our method.

\textbf{Contextual analysis:} Moreover, we investigated influence of using embedding trained on data coming from the domain source and general source. For this reason, we trained our word2vec embeddings using two datasets separately, firstly using the data dump from English Wikipedia \cite{WikiDump}, hence not including contextual information and secondly on the dataset that was built using all product descriptions, hence taking contextual approach. As it can be observed in the table \ref{similarity_per_category}, the differences in the performance are not significantly large. From the practical perspective however, the Wikipedia dataset is substantially larger, which influences both the training time and the size of the embedding. For this reason, we decided to use the embedding trained on the dataset of product descriptions in the final model.
\begin{table}
\caption{Mean similarity per text query category with and without context information available during the training stage.}
\begin{center}\begin{tabular}{c|c|c}
\setlength{\tabcolsep}{80pt}
\def\arraystretch{1.3}
Text query category & Avg similarity & Avg similarity \tabularnewline
& (Wikipedia) &  (product descriptions) \tabularnewline
& \it no context & \it with context \tabularnewline
\midrule
Color & 0.2888 & 0.2898 \tabularnewline
Human body &  0.2911 & 0.2934 \tabularnewline
Fabrics & 0.2909 & 0.2935 \tabularnewline
Style & 0.2893 & 0.2941 \tabularnewline
\bottomrule
\end{tabular}
\end{center}
\label{similarity_per_category}
\end{table}

\begin{table}
\caption{Mean number of distinct categories present in the results list for different fashion search methods}
\begin{center}\begin{tabular}{c|c}
\setlength{\tabcolsep}{80pt}
\def\arraystretch{1.3}
Method & Avg number of categories \tabularnewline

\midrule
VSE-Resnet \cite{vse} &2.47  \tabularnewline
VSE-VGG \cite{vse} &  2.87 \tabularnewline
DeepStyle-Siamese & 3.01  \tabularnewline
Late-fusion Blending & 3.08  \tabularnewline
MUTAN \cite{mutan} & 3.47 \tabularnewline
Early-fusion Blending & 3.89
\tabularnewline
\bottomrule
\end{tabular}
\end{center}
\label{result_categories}
\end{table}

\textbf{Hyperparameter analysis:} We analyze influence of hyper-parameters on blending methods. The number of final results presented to the user is set to 4 for all methods and baselines, hence we set $n_3 = 4$. Figure 12. displays how different values of $n_1$ and $n_2$ impact similarity. In the range of considered values we observe that the right balance between parameters gives similarity values higher by $0.05$ and the optimal parameters are $n_1=3$ and $n_2=4$.

\begin{figure}[h!]
\begin{center}
\includegraphics[width=0.5\textwidth]{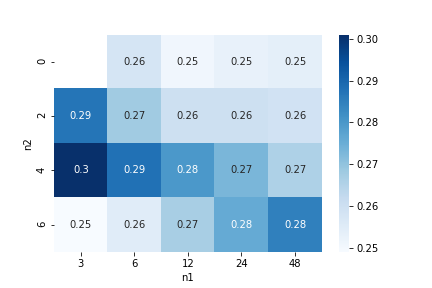}
\caption{Hyperparameter analysis for early-fusion blending and $n_3=4$ number of final results. The choice of $n_1=3$ and $n_2 = 4$ gives optimal similarity results.}
\label{n1_n2_heatmap}
\end{center}
\end{figure}

\textbf{Category diversity analysis:}
We analyze mean number of distinct object categories present in the results set (for each query and $k=4$ result items). Mean similarity as it is defined in Section \ref{sec:evaluation} depends on both name similarity as well as item co-occurence in outfit. Hence, method that would only return similar objects of the same class would not maximize the similarity metric. 
We see from Table \ref{result_categories} that VSE-Resnet has the lowest average number of distinct categories, which suggests that results from this method mostly focus on visual similarity.
On the other side of the spectrum, MUTAN \cite{mutan} and Early-Fusion Blending results have the most intra-results category diversity which means lower similarity in terms of object categories.

{\textbf{Qualitative results:}}
Fig.~\ref{semantics} and \ref{blended_results} display sample results for user queries in both fashion and interior design domains. Fig. \ref{semantics} illustrates that semantics are preserved with multimodal query and user is presented with results that combine both visual and textual queries. In the Fig. \ref{multimodal_results} detailed qualitative analysis is presented. Multimodal search results are shown for sample images with typical fashion text queries. We can see that our method is capable of retrieving visually similar results that correspond to text query but can also extend to objects from different categories that fit the semantics and have higher outfit compatibility.

\section{Web Application}
\label{sec:application}

\begin{figure}[h!]
\begin{center}
\includegraphics[width=0.5\textwidth]{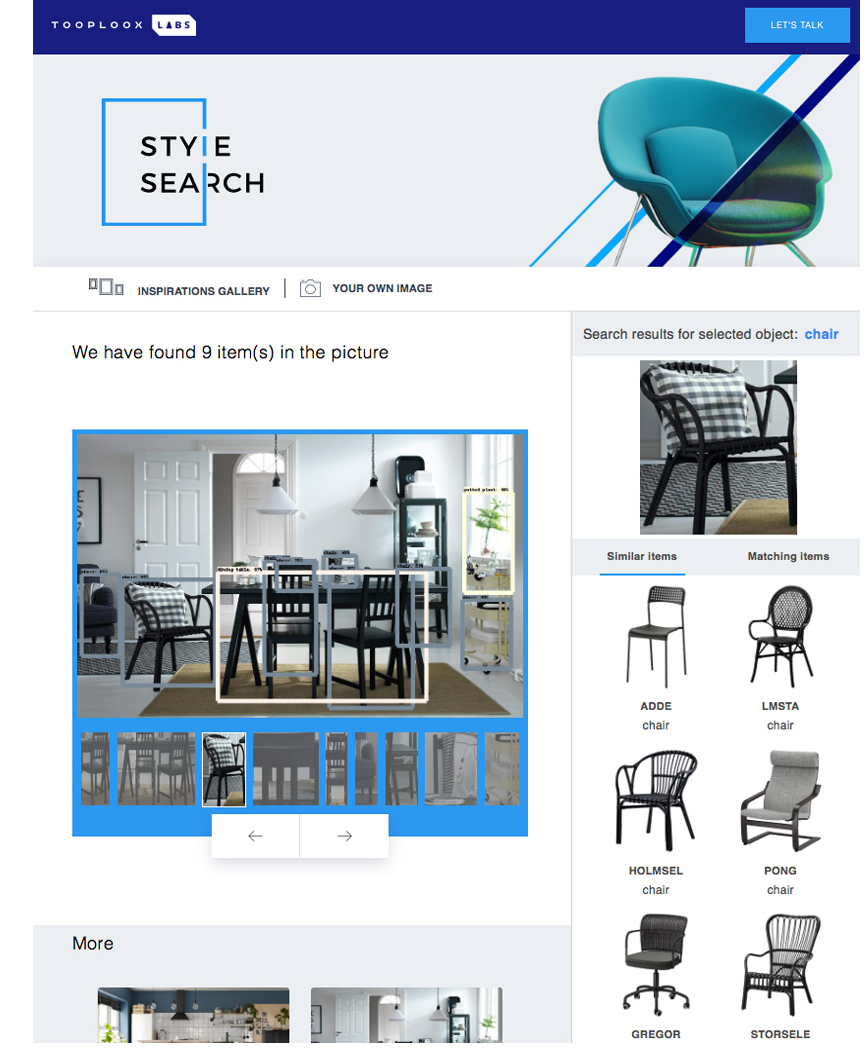}
\caption{Sample screenshot of our Style Search Engine for interior design applied in web application showing product detection and retrieval of visually similar products.}

\label{blended_results}
\end{center}
\end{figure} 

To apply our method in real-life application, we implemented a Web-based application of our Style Search Engine with application to Interior Design. The application allows the user either to choose the query image from a pre-defined set of room images or to upload his/her own image. The application was implemented using Python Flask\footnote{http://flask.pocoo.org/} - a lightweight server library. It is currently released to public\footnote{http://stylesearch.tooploox.com/}. Fig.~\ref{blended_results} shows a screenshot from the working Web application with Style Search Engine.

\section{CONCLUSIONS}
\label{sec:conclusions}
In this paper, we experiment with several different architectures for multimodal query item retrieval. This includes retrieval result blending approaches as well as joint systems, where we learn common embeddings using classification and contrastive loss functions. Our method achieves state-of-the-art results for the generation of stylistically compatible item sets using multimodal queries. We also show that our methodology can be applied to various commercial domain applications, easily adopting new e-commerce datasets by exploiting
the product images and their associated
metadata. Finally, we deploy a publicly available web implementation of our solution and release the new dataset with the IKEA furniture items.

\begin{figure*}[t!]
\centering
\includegraphics[width=1.0\textwidth]{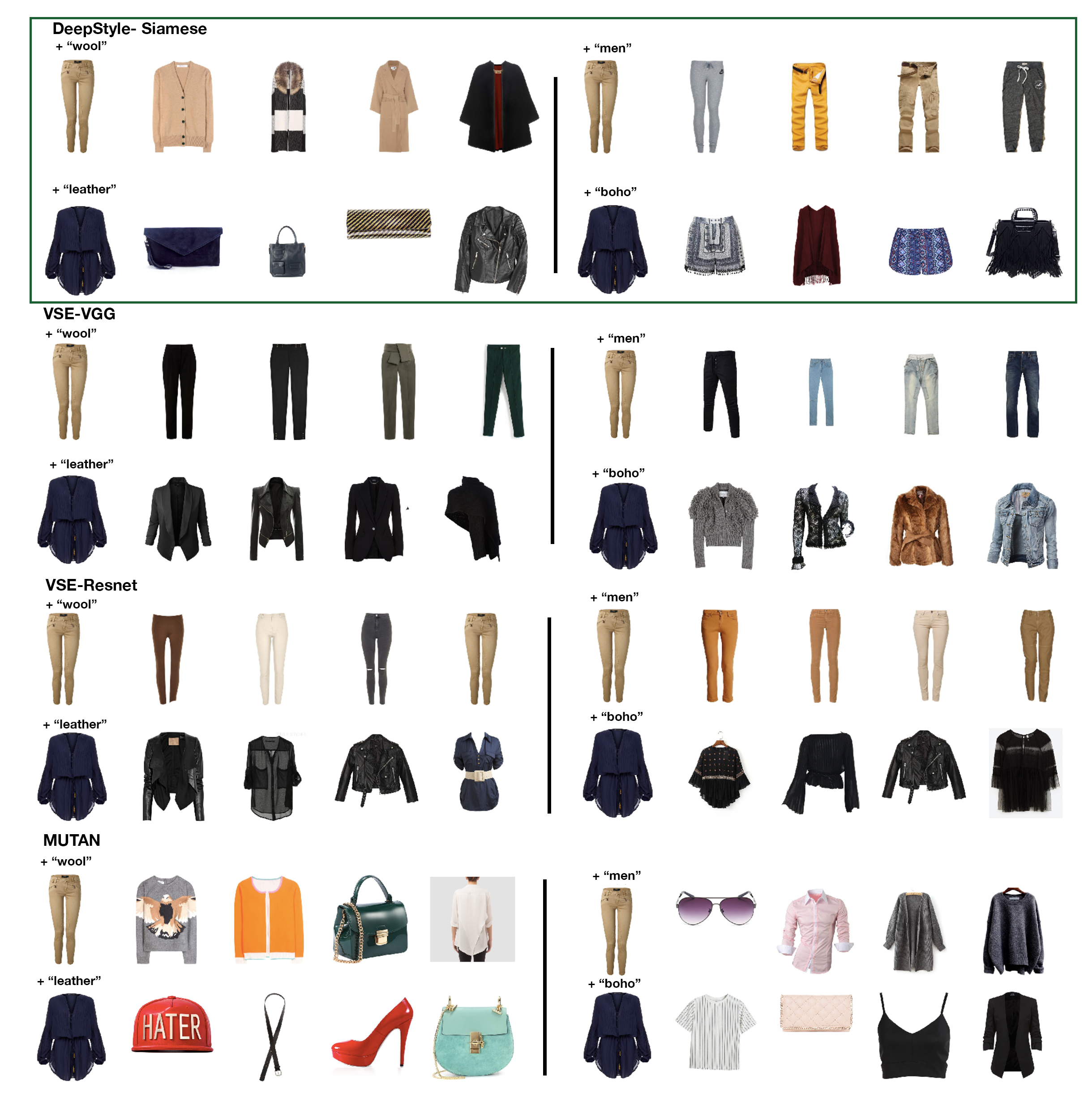} 
\caption{Sample multimodal results compared with baselines. We observe how results differ for typical fashion text queries ("wool", "men", "leather", "boho") on two test images. Our method (DeepStyle-Siamese) returns similar items corresponding to text query as well as extends visual search with items from different categories that share style similarities. We can see how our approach is different from other methods that focus mostly on shapes and colours similarity (VSE-VGG, VSE-Resnet) or methods that rely stronger on text retrieval (MUTAN).}
\label{multimodal_results}
\end{figure*}



\bibliographystyle{ieeetr}
\bibliography{bibliography.bib}

\EOD
\end{document}